# Revisiting the Exit from Nuclear Energy in Germany with NLP

Sebastian Haunss[1], André Blessing[2]

1 University of Bremen, sebastian.haunss@uni-bremen.de

2 University of Stuttgart, andre.blessing@ims.uni-stuttgart.de

## Abstract

*Annotation of political discourse is resource-intensive, but recent developments in NLP promise to automate complex annotation tasks. Fine-tuned transformer-based models outperform human annotators in some annotation tasks, but they require large manually annotated training datasets. In our contribution, we explore to which degree a manually annotated dataset can be automatically replicated with today's NLP methods, using unsupervised machine learning and zero- and few-shot learning.*

## 1. Introduction

Fine-grained annotation of political discourse is time- and labour-intensive. Therefore, most research on political discourse has so far either focused on limited cases or has switched to more coarse-grained methods of analysis like e.g. topic modelling or sentiment analysis. But recent developments in Machine Learning (ML) and Natural Language Processing (NLP) promise to (partly) automate complex annotation tasks, and thus create the opportunity for broader and more encompassing analyses based on fine-grained annotation of individual statements. Researchers have used transformer and other supervised machine learning models to predict party positions (Nikolaev, Ceron, and Padó 2023), emotions in political texts (Widmann and Wich 2023), stances in Tweets (Pattison et al. 2023), classify protest events (Bekker 2022, Wiedemann et al. 2022), or assist with the annotation of data for discourse network analysis of political claim-making (Haunss et al. 2020).

This research shows that current transformer-based models perform at some annotation tasks equally well as human annotators, and sometimes even outperform them. But usually, this performance can only be archived with large training datasets that require manual annotation. For fine-grained political discourse annotation, machine learning can therefore so far mainly be used to speed up or even automate annotation after an initial slow start of traditional manual annotation.

In the following article we test, whether this limitation can be overcome. In line with the overall perspective of this special issue, we explore to which degree the manual annotation on which Haunss, Diez, and Nullmeier (2013) based their discourse network analysis of the German debate about the exit from nuclear energy can be automated and replicated with today's NLP methods. We will thus try to replicate the original data generation process with methods of unsupervised machine learning and evaluate how far we can get today with the transformer-based revolution of language modelling (Vaswani et al. 2017).

## 2. Political backgroud: The post-Fukushima exit from nuclear energy in Germany

The German decision to exit from nuclear energy is a rare example of a policy pivot, a 180-degree shift of policy decisions in a very short time period. In less than a year, the then conservative-liberal German government coalition reverted its own nuclear energy policy decision from September 2010 to significantly extend the running time of the existing nuclear power plants. Following the 2011 Fukushima nuclear catastrophe in Japan, an intense public debate culminated in a decision of the same conservative-liberal government to immediately shut down the eight oldest nuclear power plants and to exit from nuclear energy completely by the end of 2022.

In their article, published in 2013 in this journal, Haunss, Dietz, and Nullmeier (2013) have argued that this policy pivot can only be explained by accounting for the dynamics of the policy discourse, since explanations based on more traditional policy theories fail to explain this radical and short-term change. Later studies have confirmed this general assumption (Rinscheid 2015) and have identified distinct discursive mechanisms responsible for such an unusual radical change of policy positions (Haunss and Hollway 2023).

Haunss et al.'s argument, in a nutshell, is that the demand to phase out nuclear power was able to take hold in the discourse because the actors who supported it were (1) able to occupy central positions in the discourse network. They (2) quickly succeeded in establishing a coherent set of demands that offered bridges and connection points for many other political actors with different positions and interests. And (3) the supporters of extended, longer running times of nuclear power plants failed in both dimensions, they neither succeeded in occupying central positions in the discourse network nor did they develop a coherent set of demands. The discursive dynamic towards an exit from nuclear energy was strengthened by the fact that the opposition parties avoided a strong polarisation of the discourse by connecting their own demands to more nuclear energy critical positions of some government actors and quasi radicalising these demands. This way, oppositional actors more or less pushed government actors in the direction of a nuclear phase-out (Haunss, Dietz, and Nullmeier 2013, 313–314).

Their argumentation is based on a discourse network analysis (Leifeld 2017; Leifeld and Haunss 2012) of the public debate between March and July 2011, i.e. between the nuclear accident in Fukushima and the decision of the German Parliament to phase out nuclear energy. The discourse network analysis, in turn, is based on the careful manual annotation of newspaper articles. The annotation was time-consuming, labour-intensive, and thus expensive. In the following article, we discuss, to which degree this data-generation process that in 2012 was only partly digitised and not at all automated, can today be automated with current Natural Language Processing (NLP) tools.

## 3. From newspaper articles to discourse network data

The empirical basis for the original analysis of the political discourse about the exit from nuclear energy in Germany was all articles published in two major German national newspapers, the centre-left "Süddeutsche Zeitung (SZ)" and the conservative, right-wing "Die Welt". During the observation period (11.3.2011–31.7.2011) both newspapers published 828 articles containing the keywords "(Atom* OR AKW* OR Kernenergie*) AND (ausst* OR stilll* OR abschalt* OR Laufzeit*) NOT (waffe* or bombe)". Some of these articles appeared in the local sections of the two newspapers. As in the original study, these were excluded thereby reducing the original set of articles from 828 to 773.

Among these 773 articles 398 contain at least one claim. A claim here is defined as a purposeful communicative act in the public sphere by which an actor tries to influence a specific policy or political debate. A claim can be a verbal statement or another form of action like a protest or a political decision that articulates political demands, calls to action, proposals, or criticisms (Haunss et al. 2020, 328). An actor can support/endorse or oppose/contradict a claim. Overall, these 398 articles contain 1299 claims about nuclear energy policies in Germany (Die Welt 159; SZ 239).

Annotation of the articles was done by the researchers involved in the project and by trained student assistants who were responsible for separating the relevant (containing a claim) from the non-relevant (containing no claim) articles. Each article in the final set of 398 was annotated by two annotators. The respective second annotator was always an experienced researcher and was also responsible for creating a gold standard and flagging disagreeing annotations for later discussion. Student assistants then transferred the annotations to a database. Since the annotation of both annotators was not completely independent, no intercoder reliability scores were computed.

Unfortunately, the time necessary for the annotation was not recorded when the original study was done. But from a controlled experiment in another project, we know that in a much more sophisticated annotation environment, and with a more limited annotation task, the average annotation time per article was about 10 minutes for trained annotators (Blessing et al. 2019; Haunss et al. 2020). Outside the experiment and with a task more similar to the annotation task in the nuclear policy project, average annotation time per article was more in the range of 20 minutes.

When we assume that preselecting the relevant articles takes less time than actual annotation and when we discount for all the time necessary for training the annotators and discussing the cases in which annotators disagree, a very optimistic estimation of the pure annotation time necessary for creating the discourse network dataset for the Haunss et al. (2013) article is more than 400 hours (preselection = 828 articles x 10 minutes = 138 hours; annotation = 398 articles x 20 minutes x 2 annotators = 265:20 hours). This immediately shows why discourse network analyses or political claims analyses for longer-lasting debates or with a comparative perspective are often only possible in relatively large and well-funded projects, and even there, researchers often resort to sampling strategies to reduce the amount of data that needs to be analysed (e.g. van der Brug et al. 2015; Koopmans and Statham 2000; Nullmeier et al. 2014; Wallaschek, Starke, and Brüning 2020). But sampling, of course, is problematic if we are interested in discursive interactions and dynamic processes.

A pipeline for the automatic annotation of discourse (network) data would thus open up new possibilities for research, focusing on longer time periods and larger text datasets. Such a pipeline would have to accomplish the following tasks:

- Task 1: Identify relevant articles,
- Task 2: Identify sentences/text segments that contain claims,
- Task 3: Identify actors who make the claims,
- Task 4: Classify the detected claims, i.e. assign them to the claim categories in the codebook,
- Task 5: Determine actor positions towards the respective claim.

In more technical terms, such a pipeline would need to be able to extract and classify directed, valued actor-claim dyads as shown in Figure 1 from thematically relevant texts. In such a dyad, a positive edge value denotes support, a negative edge value opposition to the respective claim.

*Figure 1: Valued actor-claim dyad*

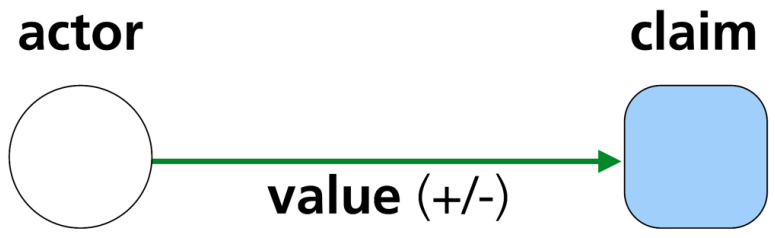

For these five steps we employ different modelling approaches. In the remaining article we discuss for each step which approach we have chosen and evaluate the automatically obtained results with the “gold standard” of the original manually annotated data set. After evaluating each step, we then compare the discourse networks obtained in a fully automatic setting with the networks from the original manually annotated dataset and discuss the potential and also the limits of current NLP tools to automate the annotation of discourse datasets.

## 4. A pipeline for the automatic annotation of political claims

The pipeline we propose here essentially replicates the steps that human annotators would do to get from a corpus of newspaper articles selected via a keyword search to a final dataset for a discourse network analysis. The pipeline is completely modular. Each task starts from a predefined state of the annotation and produces a clearly defined output. This means that each module can be replaced by an equivalent module using a different technique. In a real-world application, our pipeline would thus allow mixing human annotation, unsupervised and supervised machine annotation, or replacing one module with an equivalent module in case of technological/methodological advances in NLP modelling. Figure 2 gives a schematic representation of our annotation pipeline. The advantage of such a modularized pipeline is smaller individual tasks and more flexibility. The downside is that errors accumulate along the pipeline, so that small errors in each individual step can in the worst case add up to a significantly larger error for the whole model.

*Figure 2: Automatic annotation pipeline*

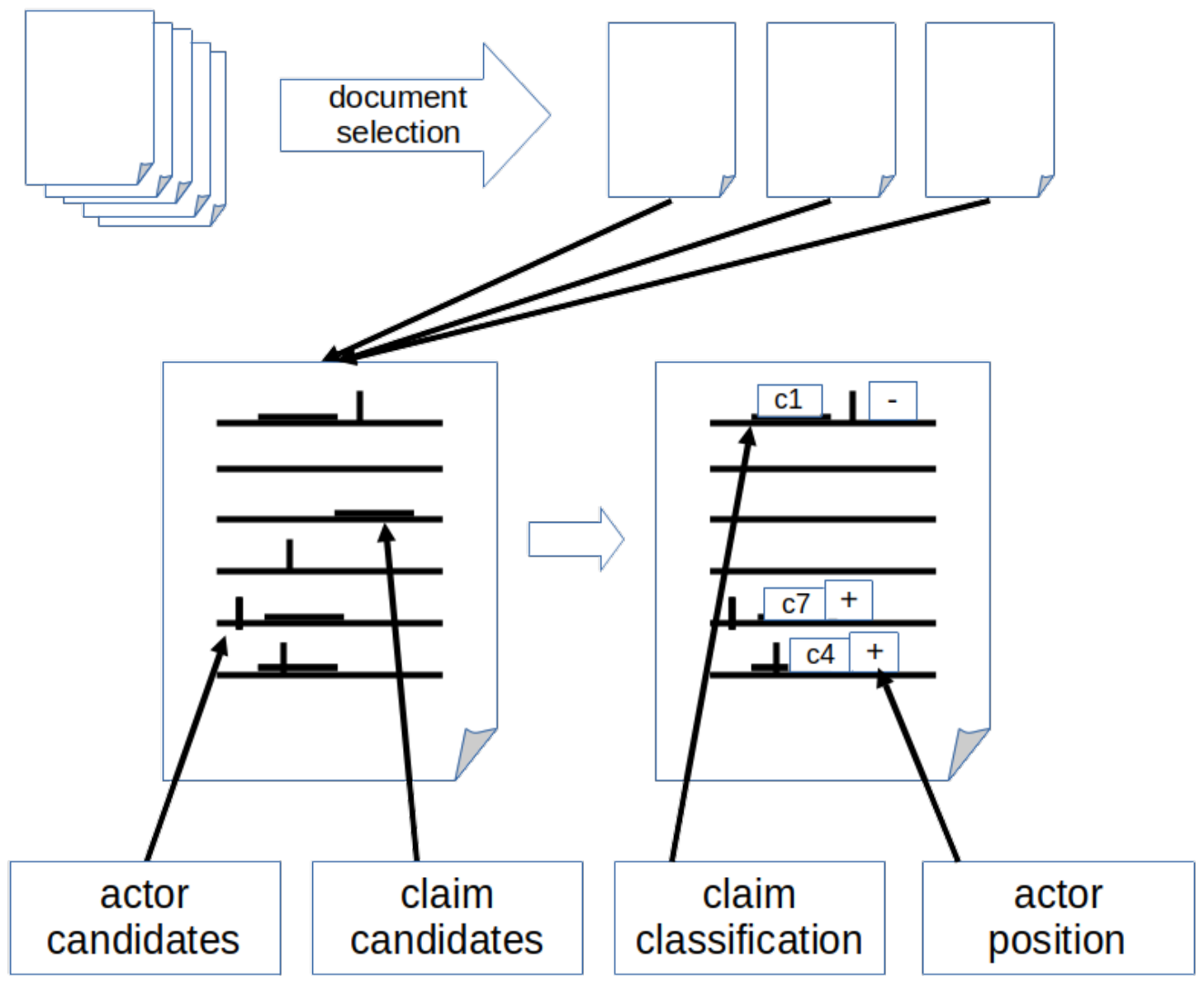

Before going into details about the modelling of the individual steps, three general caveats should be discussed: First, the manually obtained “gold standard” is slightly less golden than the term suggests. When checking the “errors” of our automatic pipeline, we occasionally encountered cases, for which the error may actually be with the human annotation and not with the classification of the model. These human errors do not invalidate our original dataset, but they remind us that human annotation is inherently an interpretive abstraction task, and even when quality checks are implemented, different (groups of) annotators may disagree on the correct categorization of a statement. Human annotation will thus often not be perfect, and error margins of automatic systems should therefore be evaluated in relation to error margins of human annotators for the same task.

Second, annotation of the original dataset is complete from a political science perspective but incomplete from a computational linguistics perspective. The same actor-claim dyad was only annotated once per article. So, if a journalist reports in one sentence about a politician's claim and then cites the same politician with a direct quotation in the next sentence, only one of the two instances of the same actor-claim dyad was manually annotated, because the second sentence does not provide additional information, it merely presents the same actor-claim dyad in a different form. From a computer linguistics perspective this annotation is incomplete, because the fact that also the second sentence contains an actor-claim dyad was not recorded. This incomplete manual annotation creates problems when trying to assess the quality of the automatic annotation, because we cannot be sure that only manually annotated text segments may contain claims. We will discuss this in more detail in section 6.

Third, we omitted one crucial element from our automation task: the generation of a codebook. Codebooks are usually generated in an iterative process, starting from a number of theoretically derived categories or from previous knowledge from earlier research, and then refined during the initial phase of the annotation process. For this article, we took the existence of clearly defined code categories as given. This is not a completely implausible assumption. In well researched subject fields, established codebooks may exist that may only require minor adaptations for a new research project. For the debate about exit from nuclear energy in Germany this was not the case. But we nevertheless decided to omit the problem of codebook creation because (a) the problem of codebook creation is on a different conceptual level than the other five tasks (it is mainly a problem of finding the right level of abstraction for the desired analysis) and (b) a valid codebook could be generated in a limited pilot study with a small sample of manually annotated articles, leaving still the main annotation work to computers.

In the following part we discuss how we approach each of the five tasks. We try to give a general description that should be comprehensible also for readers not familiar with computational language modelling. If necessary, we provide more technical details for readers interested in replicating our claim detection and classification pipeline towards the end of the respective section.

### 4.1. Task 1: Identifying relevant articles

Initial article selection is usually done via keyword search in newspaper databases or archives. Since keywords are sometimes ambiguous and articles containing relevant keywords may nevertheless be irrelevant because they e.g. have a different geographical focus, the total number of articles found with a keyword search is usually higher than the number of relevant articles. In some contexts, the ratio of found to relevant articles is in the region of 10:1 or even higher. The relative specificity of the keywords relevant to exit from nuclear energy in Germany after Fukushima resulted in only 2:1 found vs. relevant articles. For the current analysis we decided to not filter out irrelevant articles, since they either should not contain a claim, anyhow, and thus should drop out in the second step of our pipeline. Or they may contain claims about nuclear energy policies in countries outside Germany. In this case we expect to catch most of the irrelevant claims when looking at the actors who make the claims.

### 4.2. Task 2: Identify sentences/text segments that contain claims

Preparatory steps for tasks 2 to 5 are downloading the article fulltexts from Factiva using the article IDs. Annotated text segments from the original study are then mapped to their respective articles in order to get character positions of their start and end points. Each article is pre-processed by using NLP (Natural Language Processing) tools. We use spaCy (version 3.6, de_dep_news_trf) for sentence splitting, tokenization, part-of-speech-tagging, noun chunking, and dependency parsing. For named entity recognition several models exist for German. After manual evaluation of some sample sentences of our domain data we decided to use the FLAIR-NER model (NER German large, https://huggingface.co/flair/ner-german-large; Yu, Bohnet, and Poesio 2020).

We implement task 2 as a binary sentence classifier that classifies each sentence either as a claim candidate or as containing no claim. The sentence classifier represents each sentence as a 768-dimensional vector – a so-called embedding – by using a sentence transformer model for the encoding (sentence-BERT, paraphrase-multilingual-mpnet-base-v2; Reimers and Gurevych 2019). We can think of these embedding vectors as a coordinate in a high-dimensional space that locates more similar sentences closer to each other than to sentences with a different meaning. The embeddings vector thus incorporates linguistic and contextual features to assess the potential claim status accurately.

After computing the embeddings, we then use a neural network (MLP) binary classifier that distinguishes between claim and non-claim sentences. This classifier had been trained on the DEbateNet-mig15 dataset containing 556.185 tokens from 959 newspaper articles about refugees and migration from the German daily newspaper "taz – die tageszeigung" which had been manually annotated in a separate project (Lapesa et al. 2020; Padó et al. 2019).

Even though the classifier was originally developed to identify migration politics related claims in German newspapers, the resulting model seems to be relatively topic agnostic, so that it can also be used to predict claims in other issue areas as well. We believe that our binary claim classifier picks up general semantic and syntactic features of sentences expressing claims, that also enable humans to identify claim sentences independent of the specific topic, e.g. the presence of active verbs like demand, want, criticize, announce, warn, etc.

Technically the classifier assigns to each sentence a score between 0 and 1, representing the likelihood of it being a claim. To optimise the recall, we strategically set the threshold for claim candidacy determination at a low value (0.1). This ensures that a broader range of sentences with reasonably high scores are considered as claim candidates in the initial phase. In order to enhance the precision of the identified claim candidates, we implement in a later step a filtering process. The filtering criteria are designed to eliminate false positives and retain only the most relevant and contextually valid claim candidates.

### 4.3 Task 3: Identify actors who make the claims

Recognizing actors in textual data plays a pivotal role in claim identification, as it helps establish the source and context of claims. An actor (a person or an institution/organisation) is always the anchor for a possible claim. A claim always requires an actor who makes the claim. In this study, we propose a comprehensive approach to identify actors in sentences, which are then linked to corresponding claim candidates. Initially, we employ Named Entity Recognition (NER) to identify and extract mentions of persons and organisations from the articles. This step allows us to isolate potential actor candidates in the text. The automatically recognized named entities are the major source for the actor candidates. Additionally, we use the POS-tags to identify pronouns and utilise dependency parsing to analyse the syntactic structure of sentences and identify the verbs associated with the actor candidates. Together this allows us to determine whether a sentence represents an inside or an outside case: In the inside-case, the actor is explicitly mentioned within a sentence, often accompanied by verbs that imply direct action or expression, such as "fordern" (demand) or "plädieren" (plead). In the outside-case, the claim sentence uses pronouns like "wir" (we), "ich" (I), or “er/sie” (he/she, him/her) as the subject, indicating an actor outside the sentence context. In this scenario, verbs like "sagte" (said), "betonte" (emphasized), or "kritisierte" (criticized) are typically used. Later we normalise actor names by trying to find for each actor the longest match in the whole dataset and check for possible mismatches, which could be caused by different spellings (e.g. only the lastname or abbreviations), or the German genitive “s”.

The final step involves matching the identified actor candidates with their respective claim candidates based on their associated verbs. Only claim candidates with an appropriately linked actor are considered for further processing.

For the present analysis we focus on the inside-case, self-contained claims with claim and named actor in the same sentence, only. On the one hand, this is based on the assumption that newspaper articles usually contain redundant information. The same actor-claim dyad is often reported more than once within one article. For a correct result, the model only needs to find one occurrence of each distinct dyad within each article. The coreference resolution – linking pronouns to the respective named actor – may therefore often not be necessary to still capture the relevant information. On the other hand, our decision to use Sentence-BERT for the claim classifications, limits the context awareness of our models to individual sentences only. We therefore cannot reliably predict claims that span more than one sentence and therefore cannot know whether an actor mentioned in a preceding sentence may “belong” to a claim in a later sentence.

In principle, integrating coreference resolution using e.g. a neural network approach as described by Schröder, Hatzel, and Biemann (2021), would be possible, but would require additional changes in modelling the claims.

### 4.4 Task 4: Classify the found claims

For the classification we start with our codebook categories. As mentioned above, we assume that a reasonable codebook exists for the topic at hand. In principle, computational models could be used to generate possible code categories (Zhang et al. 2022). But since finding the “right” code categories is ultimately a decision of finding the right level of abstraction for a desired analysis, computational tools for codebook creation go beyond the scope of this article.

To begin with, we transform the category labels into sentence embeddings. Each sentence and each category label is embedded using a S-BERT-model (paraphrase-multilingual-mpnet-base-v2). This conversion facilitates the comparison of categories with claim candidates in an embedding space, enabling the identification of semantically related categories. Using the sentence embeddings of categories, we identify for each category in the codebook sentences that serve as seeds for claim categorization. This selection of anchor sentences for each codebook category is the only manual annotation step in our pipeline presented here. Figure 3 shows on the left panel the code categories used in the analysis and on the right panel an example for text spans the model suggests as most similar to the code category “re-evaluate nuclear power policies (Atompolitik auf den Prüfstand)”. Seed sentences are manually selected from text spans suggested by the model. These seed sentences act as representatives of their respective categories and play a crucial role in the subsequent similarity comparisons.

*Figure 3: Code categories and seed sentences*

**100 Ausstieg**
101 Ausstieg (schnell)
102 Ausstieg (sofort)
103 Ausstieg (endgültig)
104 Ausstieg (mit Deadline)
105 Abschalten der Altmeiler
106 endgültige Stilllegung der Altmeiler
110 Moratorium
120 Atom-Politik auf dem Prüfstand
130 Laufzeitverlängerung
140 Neubau von AKWs
150 Atomkraft als Brückentechnologie

**200 Sicherheit**
201 Sicherheitsüberprüfung
202 Stresstest
203 Stresstest (EU-weit)
205 Nachrüstung
206 Sanierung der Zwischenlager

**300 Energiepolitik**
301 Energiewende
302 Investition in erneuerbare Energie
305 Energieeffizienz
310 Stromnetzausbau
311 Gaskraftwerke
312 Kohlekraftwerke

**400 Verfahren**
401 Alleingang
402 Übereiltes Handeln
403 Konsens/Einigkeit
405 Populismus
410 Schadensersatz
420 gesetzliche Basis schaffen
425 EU-weite Regelungen
430 Deutschland als Vorbild
440 Verantwortungsbewusst-sein
450 Wirtschaftlichkeit

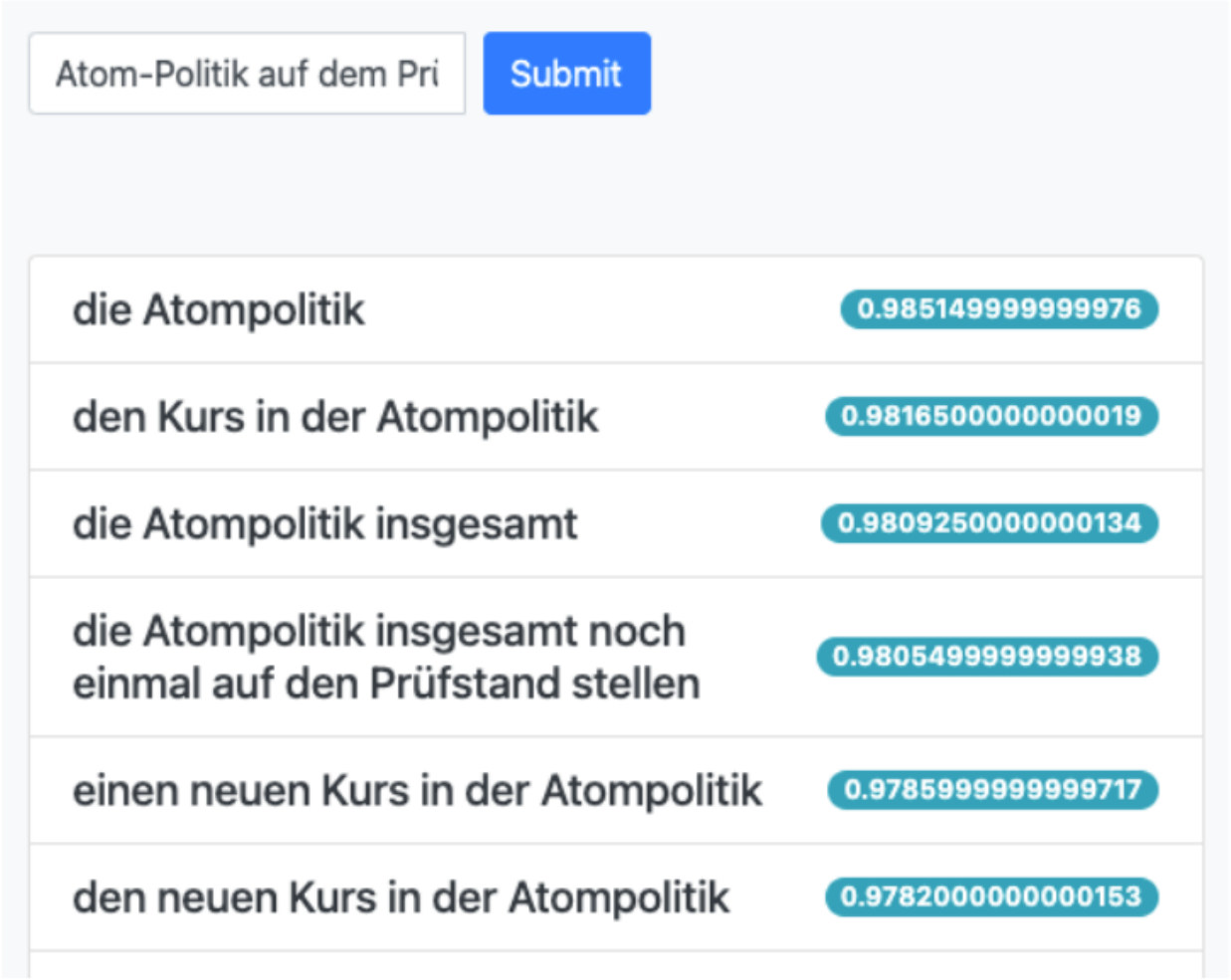

For the classification of the candidate claims we then perform pairwise comparisons between the embeddings of each claim candidate and all seed embeddings by computing their cosine similarity. The similarity score obtained from each comparison represents the degree of similarity between the claim candidate and each category seed. Based on the similarity scores, we categorise each claim candidate by assigning it to the most similar category. This process ensures that each claim candidate is associated with the category that exhibits the highest semantic similarity.

To control the precision of claim categorization, we introduce a threshold for similarity scores. Claim candidates with similarity scores above this threshold are considered relevant and retained, while those with scores below the threshold are filtered out as potentially irrelevant.

It should be noted that in the original analysis all claims annotated with the categories "other" and with the general "procedures" category were deleted from the original dataset (= 25 actor-claim dyads) because the claims annotated with these categories were so diverse that their inclusion in the discourse networks would have created artificial connections between actors who de facto did not agree on the same claim. In line with this we did not include example sentences for these categories in the training dataset for our classifier.

### 4.5 Task 5: Determine actor positions towards the respective claim.

In order to ascertain the positions of various actors towards the respective claim, we employ a NLI (Natural Language Inference) based zero-shot classifier. NLI, also known as recognizing textual entailment, is a fundamental task in NLP aimed at evaluating the semantic relationship between two given text fragments. The goal of NLI is to determine whether a hypothesis can be logically inferred from a given premise. This task holds significant importance as it addresses the challenge of understanding and modelling the intricate nuances of human language comprehension and reasoning. In our scenario we compare each claim against a positively formulated version of the claim category and a negatively formulated claim category. The pair with the higher score has a higher entailment and indicates the position of the actor. To illustrate this process, let's consider an example: the claim sentence "Angela Merkel calls for a swift shutdown of all nuclear power plants." The associated claim category is "swift nuclear phase-out." Now, the claim sentence is compared once with the supporting phrase "swift nuclear phase-out" and once with the opposing phrase "against a swift nuclear phase-out." In this case, the first pair is expected to receive a higher score, indicating a more favourable stance towards the "swift nuclear phase-out" claim category. This methodology enables us to systematically assess actor positions and provides valuable insights into the discourse surrounding the claims under consideration.

## 5. Evaluating the performance of the pipeline

How does our pipeline perform on the deliberately chosen hard task of getting from keyword-selected newspaper articles to discourse networks (almost) without supervision? We first present a quick overview and then go into details for each aspect. We evaluate our automatic annotation against the manual annotation of the original dataset.

- *Task 1: Identification of relevant articles*: We decided to merge this task with the second task, expecting that the claim identifier would implicitly find the relevant articles anyway. This turned out to be problematic as we will show in section 5.1. In the outlook we will discuss how this could be addressed in future applications.
- *Task 2: Identification of sentences/text segments that contain claims:* Topic agnostic claim identification works surprisingly well. Our classifier, trained on data from a completely different topic area (migration), predicts the claims in our nuclear energy exit debate dataset quite well. The model seems to pick up some more general features of political claims and thus works quite well as a topic agnostic claim detector.
- *Task 3: Identification of actors who make the claims:* Our pipeline finds the most central actors for each time period, but overall actor identification is still not reliable enough. The core problems are that we do not have a working model for coreference resolution and therefore can only identify explicitly named actors. The second problem is journalists' tendencies to use synonyms in order to avoid repetition. E.g. an article mentioning Angela Merkel in one sentence may talk in the next sentence about the chancellor ("die Bundeskanzlerin") and then in the next sentence "the chief of government".

- *Task 4: Classification of the found claims*: With only one sentence per claim category the model predicts claim categories with a macro average precision of 0.55. This is not yet satisfactory, but as we show below, misclassifications often happen between neighbouring categories.
- *Task 5: Determination of actor positions towards the respective claim*: Our approach reaches an accuracy of 75.8 %. In more detail: 755 true positives, 102 true negatives, 142 false positives, and 132 false negatives. Given that we trained a binary positionality classifier by simply adding the negative phrase "warnt vor" (warns for) to our seed sentences, this is a very good result, but of course too simplistic and not optimal for all classes.

In the following, we discuss the performance of our pipeline in more detail and provide a more nuanced evaluation of its strengths and weaknesses.

### 5.1. Detecting relevant articles and claims

The claim classifier provides an implicit article selection as it does not find claims in all articles. The set of articles in which the classifier predicts claims contains 443 of the original 773 articles. In 290 articles claims were originally annotated and also predicted from our model. So about three quarters of the manually selected articles are also selected in this implicit article selection process, but the classifier also includes a significant number of additional articles (153) in which it also finds claims. The problem now is, that we do not know which proportion of these seemingly false positives really are false. In fact, there are three false positives scenarios in which the "false" positives are not completely false:
First, it may be that the human annotators have overlooked a claim. Second, in some cases the classifier finds sentences that actually contain claims, but claims in other issue areas than nuclear energy policy. These are real false positives, but due to the topic agnostic nature of the claim detection they are almost unavoidable by-catch. And third, in some cases the classifier correctly identifies claims in sentences that have not been annotated because the same claim-actor dyad already appeared in another article.
This last case points again to the information redundancy issue in newspaper articles: When manually annotating texts, human coders responsible for the original dataset skipped reoccurrences of the same claim by the same actor after its first occurrence in a given article and across articles. As discussed above this is reasonable in a manual setting from a social sciences perspective because repeated actor-claim dyads on the same date do not contain additional substantial information. Nevertheless, this sparse – and from a computer linguistics perspective incomplete – annotation practice biased model metrics towards inflated false positives and false negatives, both on the article and on the sentence level, because to correctly identify an actor-claim dyad it doesn't matter in which sentence of an article or in which article on the same date the model finds this dyad. If the model only finds the second occurrence, then the actual result is still correct, but model metrics would record a false negative for a claim not found at the first occurrence and a false positive for a claim found in the second occurrence. In fact, in our case the model finds the correct actor-claim dyad in more than 50 percent of the cases in a different sentence than the human annotators.
The first case reminds us that even a large carefully curated gold standard annotation may still contain some errors. And the second case is a result of the model's ability to find claims independent of the topic of the claim. In one of the articles the model e.g. classifies a sentence in which two members of the parliament call for allowing preimplantation genetic diagnosis in a strictly limited framework as a claim for "operating time extension". The sentence clearly contains a claim, and given the fact that the model can only choose among the topic specific categories, the choice is not completely implausible. But the resulting actor-claim dyad is nevertheless wrong.
How well do the claims and actors found by our model represent the actor-claim dyads in the manually annotated text corpus? To answer this question, we now compare the temporal and topical distribution of the claims predicted by our model and those manually annotated in the original dataset.

## 5.2. Frequency distribution of claims

Figure 4 plots the weekly sums of the number of distinct actor-claim dyads per day found by the classifier and in the manual annotation. In all but three weeks, the model predicts more actor-claim dyads than the manual annotation, but the distribution of actor-claim dyads over time seems to reflect the distribution from the original analysis quite well.

*Figure 4: Distribution of claims over time*

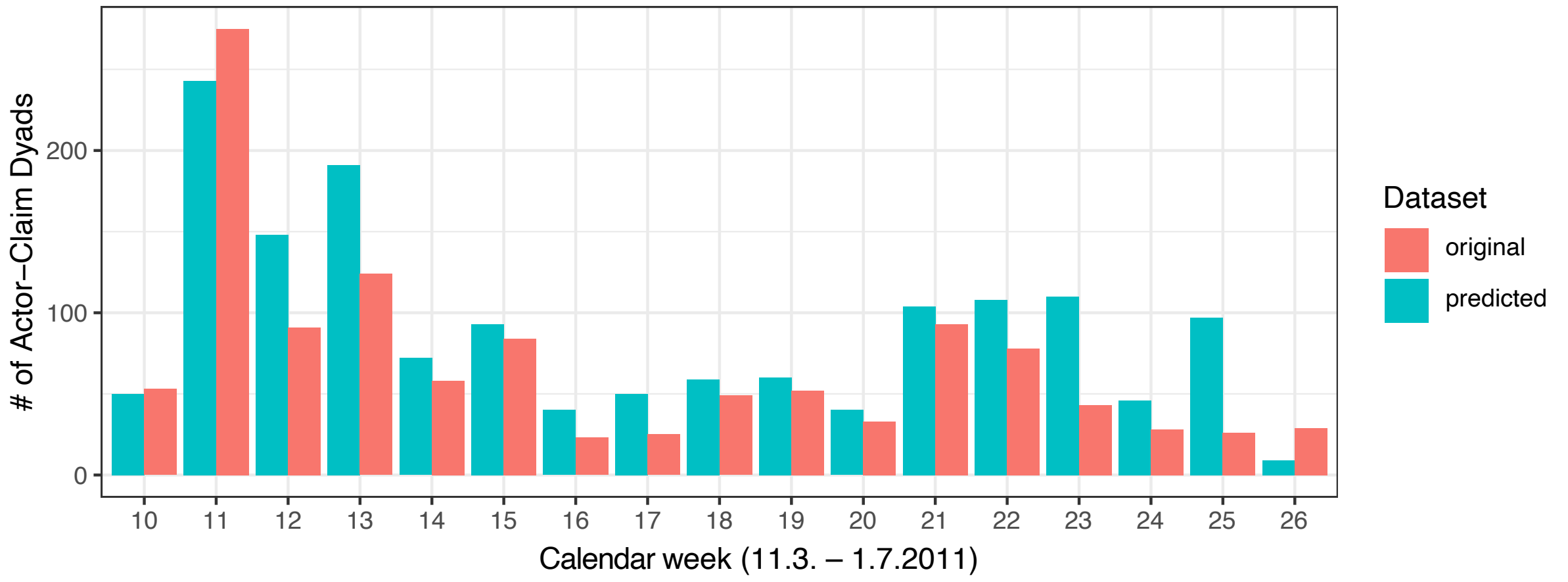

A closer inspection of the predicted claim categories delivers a slightly less optimistic image. Looking at individual categories and at shorter time spans, we can see in Figure 5 that the accuracy of the model fluctuates strongly across observation periods. In week 11 the model is much closer to the manual annotation than in week 20. We observe a general pattern that in weeks with higher numbers of claims, i.e. in weeks with more intense debates, the fit between the automatically predicted the manual annotation is better than in weeks with fewer claims.

*Figure 5: Claim categories in selected weeks*

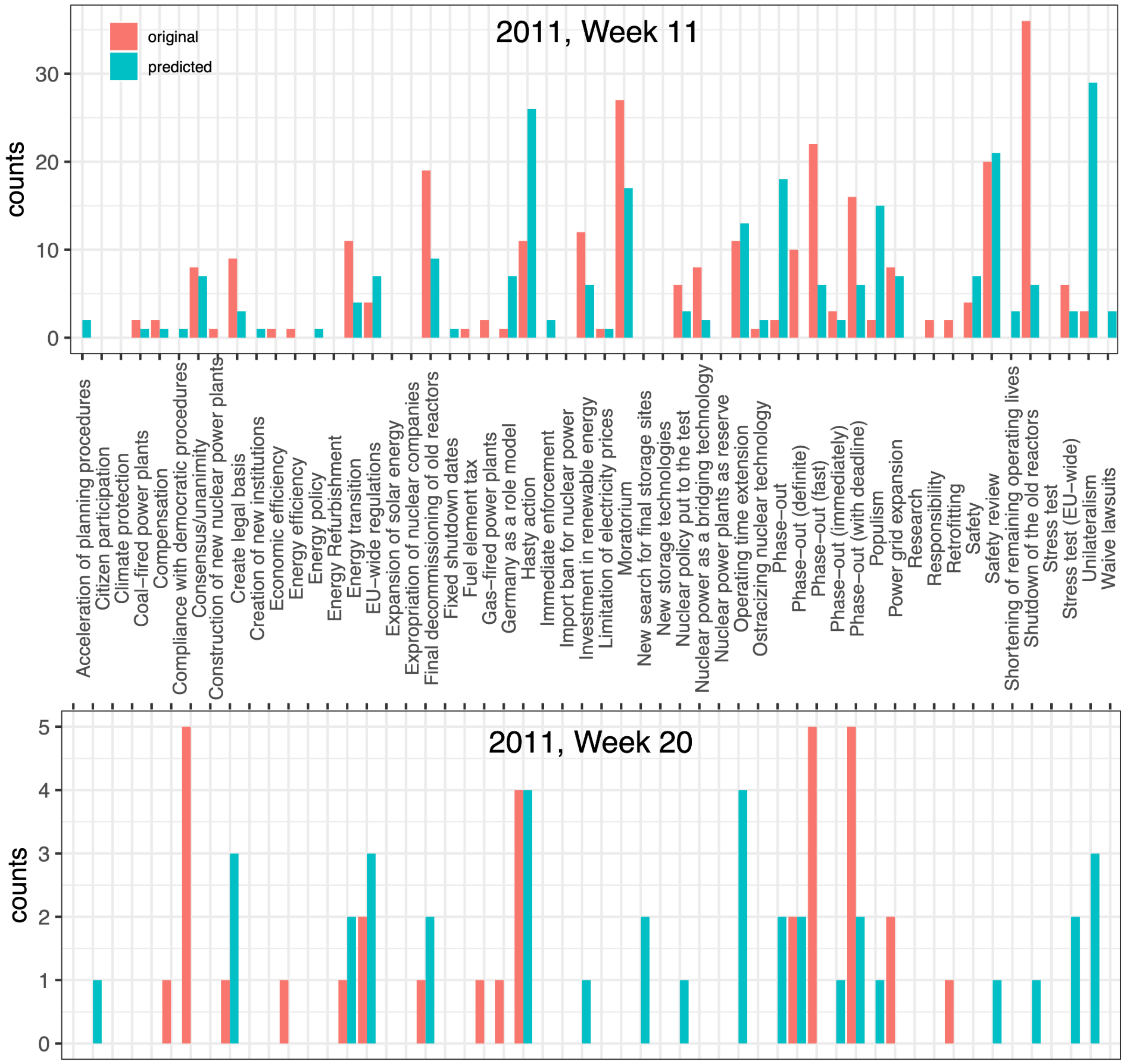

## 5.3. Claim categories

The bar charts do not differentiate between false positives (claim found even though there was no claim) and misclassifications (claim found but not assigned to the correct category). In order to evaluate the classification errors, we now let the model classify all claims from the manual annotation. We thus no longer include the false positives and false negatives but assess only the errors the model makes on a given set of known claims. Figure 6 provides a confusion matrix of the gold-standard (manually annotated) vs. predicted categories. In an ideal scenario the confusion matrix would only show entries in the main diagonal. In the real world we see that, indeed, most entries are in the main diagonal, the model thus classifies the majority of the claims correctly.

*Figure 6: Confusion matrix actual vs. predicted claims*

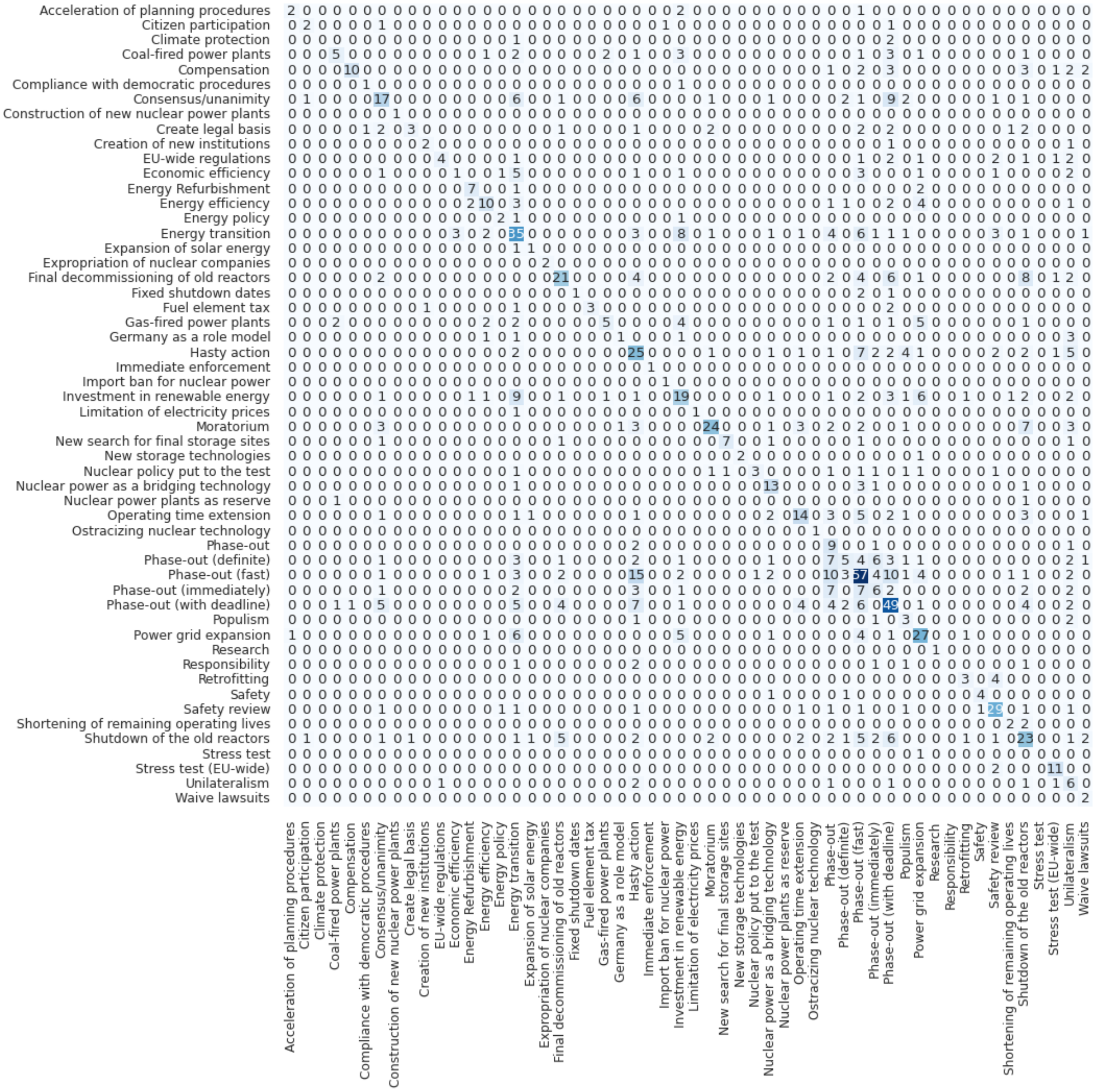

However, there are also many entries outside the main diagonal. In particular we see a cluster of entries around the category “Phase-out (fast)”. Here the model has misclassified some claims in other phase-out categories. Obviously, the model has a hard time dealing with the very fine-grained differentiation between different phase-out claims – a problem it shares with human annotators. Another recurring classification error is between “Energy transition” and "Investment in renewable energy". Again, the model is not completely off the mark and confuses categories that are on a content level close neighbours. Also, the model has classified some of the "Phase-out (fast)" claims as "Hasty action", a claim that was used by actors in favour of nuclear energy, warning against hasty action. The interesting thing is that when we look at the language used in these claims this misclassification is understandable, since both claims address an acceleration of political decision making – fast exit, hasty action. The fact that misclassification errors are often between closely related categories is an indicator of the high level of semantic understanding the model delivers.

At first sight this misclassification into similar categories error seems to point to a problem with not well enough separated categories in the original codebook. But the many phase-out categories were actually a consequence of the development of the political discourse, in which some actors e.g. made claims against "fast" and for "immediate" phase-out. Not separating these two seemingly similar demands would have created in the network false links between actors that actually opposed each other. The problem thus goes further than poor category specification in the codebook.

## 5.4. Discourse Networks

What kind of discourse networks does the automatic pipeline produce? How similar are they to the networks from the manual annotation? To evaluate this we first analyse the number of correctly and falsely identified actors and concepts in each of the eight time periods in which the overall observation period was partitioned in the original article (Haunss, Dietz, and Nullmeier 2013, 304). Table 1 reports for each of the time periods the overlap between actors and claims present in the aggregated core networks from manual and automatic annotation. We report precision, recall and F1 scores, where precision is the fraction of true positives among all predictions, recall (or sensitivity) is the fraction of found relevant instances, and F1 is the harmonic mean of precision and recall. We stick to the core level that was chosen in the original article. An n-core is the network that results from deleting all nodes with a degree centrality below n. A 3-core network thus retains only those nodes with a degree of at least 3 and their corresponding edges. As in the original article we limit the deletion of nodes to the concept partition of the network and thus retain all actors connected to claims that received at least n mentions in the respective time period.

We can see that for all but the first and third periods the F1 score is much better for claims than for actors. To a large extent this is a result of the decision to create network cores based on the minimum indegree of claims and retain all actors connected to these claims. But it also shows that the seemingly simple task of actor identification is more difficult than one might expect. The F1 score for dyads is very low and it is generally decreasing over time.

*Table 1: Metrics for actors, concepts, and dyads in the eight aggregated networks*

| Time period | n-core | Actors | | | Claims | | | Dyads | | |
|---|---|---|---|---|---|---|---|---|---|---|
| | | F1 | Prec. | Recall | F1 | Prec. | Recall | F1 | Prec. | Recall |
| 1: 11.–13.3.2011 | 3 | **0.59** | 0.58 | 0.61 | **0.57** | 0.57 | 0.57 | **0.31** | 0.33 | 0.29 |
| 2: 14.–15.3.2011 | 5 | **0.34** | 0.29 | 0.41 | **0.47** | 0.36 | 0.67 | **0.18** | 0.15 | 0.22 |
| 3: 16.–22.3.2011 | 6 | **0.34** | 0.33 | 0.35 | **0.29** | 0.30 | 0.27 | **0.08** | 0.08 | 0.09 |
| 4: 23.3.–8.4.2011 | 7 | **0.36** | 0.30 | 0.47 | **0.69** | 0.60 | 0.82 | **0.15** | 0.12 | 0.18 |
| 5: 9.4.–28.4.2011 | 6 | **0.32** | 0.25 | 0.45 | **0.55** | 0.50 | 0.60 | **0.13** | 0.12 | 0.15 |
| 6: 29.4.–17.5.2011 | 6 | **0.17** | 0.13 | 0.26 | **0.42** | 0.29 | 0.71 | **0.07** | 0.05 | 0.11 |
| 7: 18.5.–30.5.2011 | 6 | **0.33** | 0.29 | 0.38 | **0.44** | 0.40 | 0.50 | **0.06** | 0.05 | 0.07 |
| 8: 31.5.–1.7.2011 | 7 | **0.28** | 0.23 | 0.36 | **0.38** | 0.31 | 0.50 | **0.08** | 0.06 | 0.12 |

When we now visually inspect the networks from two select time periods (Figure 7 and 8) we can see that the model is able to replicate some aspects of the original networks but that much information is lost and many additional incorrectly identified actor-claim dyads are added to the networks. The graphs show the networks from the manually annotated data on the left and the graphs from the automatically annotated data on the right. Actors and claims that appear in both networks are highlighted in the right network with a red border and edges that appear in both networks are highlighted in the right network with darker grey. The graphs confirm the results from Table 1 that the model picks up a large part of the actors and claims present in the discourse, but the ratio of correctly predicted dyads is unfortunately rather low.

*Figure 7: Core networks (3-core) from manual and automatic annotation, time period 1 (11.–13.3.2011)*

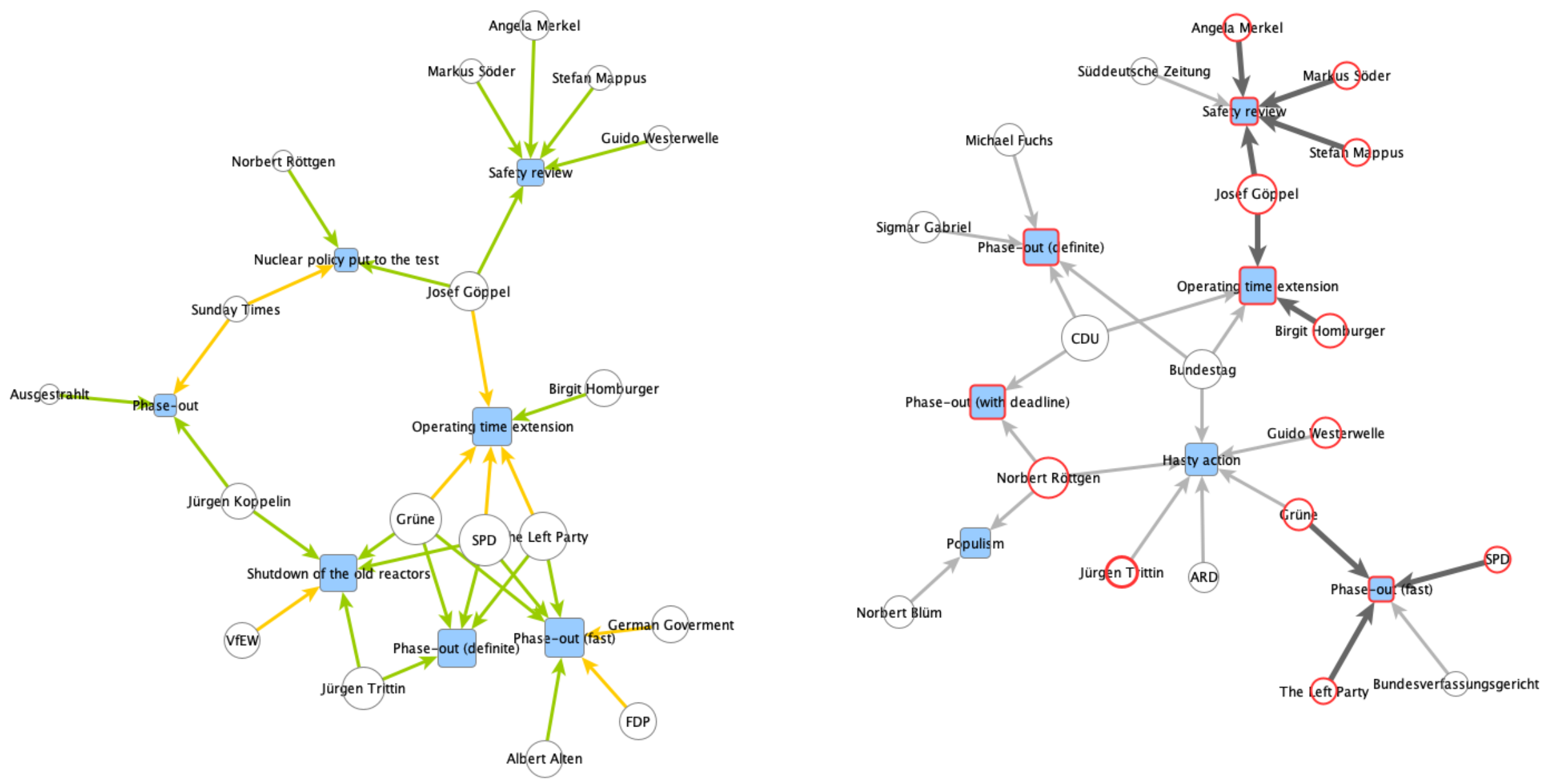

*Figure 8: Core networks (7-core) from manual and automatic annotation, time period 4 (23.3.–8.4.2011)*

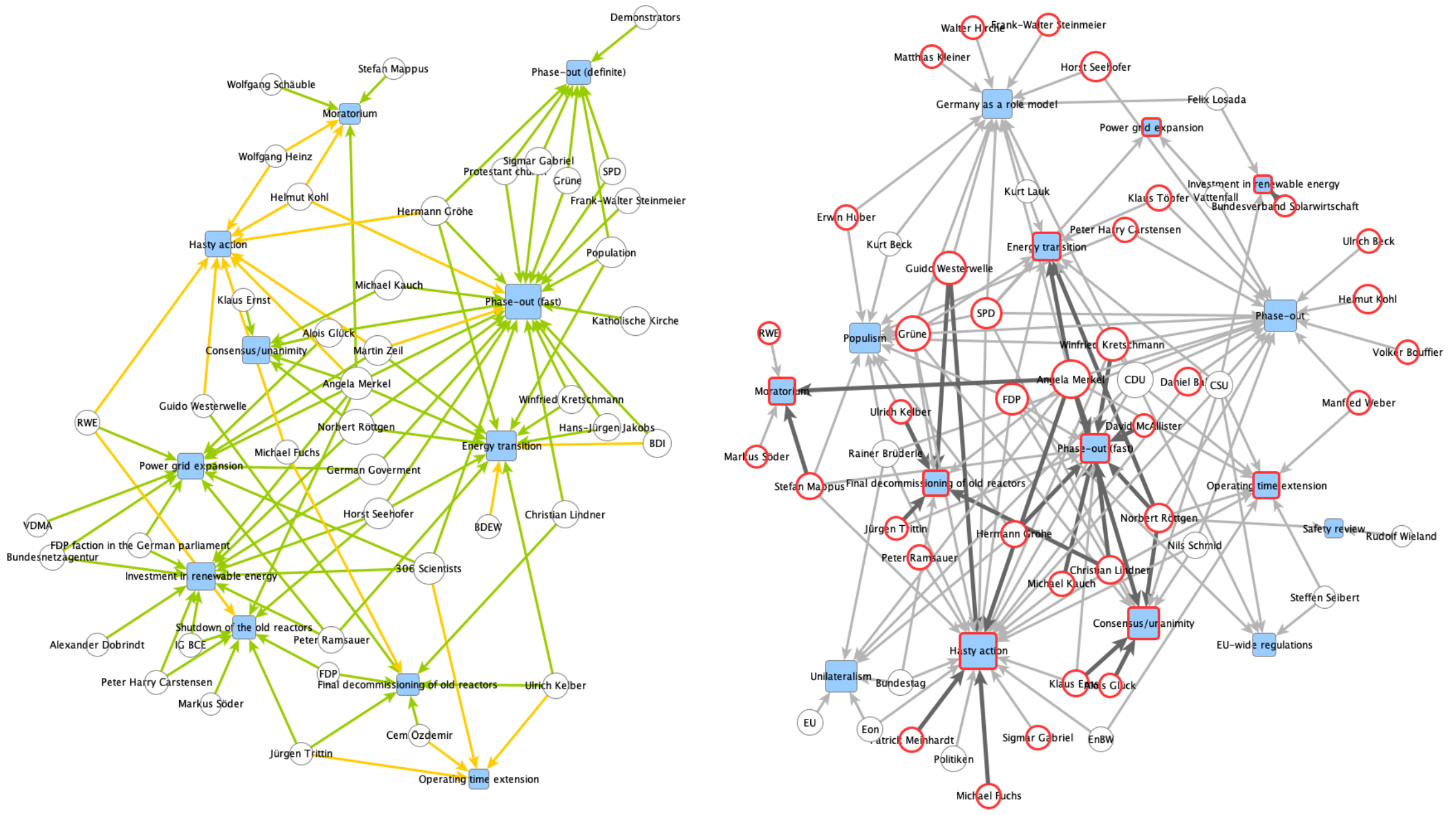

One reason for the low number of correctly predicted dyads becomes apparent in Figure 8 in which the model often predicts the unqualified "phase-out" claim, which was no longer used in the manual annotation after the first week, when the authors of the original study observed a differentiation into various competing phase-out claims.

On a substantive level the automatically generated network for the fourth period (Figure 8) correctly captures the shift towards a discussion of renewable energy. This claim was put forward already one month after Fukushima by actors from the governing coalition and we had interpreted its occurrence as an indicator that already at this early stage of the debate the decision towards an exit from nuclear energy was predetermined. The graph also captures the growing centrality of government actors during this time period. But it misrepresents the centrality of concepts by not capturing the increased centrality of the exit claims during this time period.

It should be noted that the metrics reported in Table 1 do not yet account for the actors' position towards the claim and counts a dyad as a match independent of whether it is a statement for or against the respective claim.

## 6. How far can we get with automatic discourse network analysis?

In this paper we present an attempt to use state of the art language modelling to (almost)

completely replicate a complex manual annotation pipeline for the generation of data for discourse network analysis. We thus evaluate the possibilities and limits of using current approaches from computer linguistics to perform qualitative annotation on a large newspaper dataset in the social sciences.

Ten years ago, when the original article was written, such an attempt would have been completely delusional. Today, we still cannot provide a simple pipeline that would take newspaper articles as an input and produce discourse network data as an output on which a reliable analysis of the dynamics of a given discourse may be based. As Figures 7 and 8 show, the resulting networks still contain too many errors, so that it is not yet possible to base a scientific or political analysis on them.

But we have deliberately chosen a very hard task. We expected the models to perform their classification tasks "out of the box". The only additional information that we provided was one example sentence for each claim category. We refrained from providing topic specific training data because the goal was to evaluate how far we can get without human intervention. With this maximalist approach we get results that are not yet usable without human supervision. We now discuss for each step what could be done to improve the predictions while still keeping supervision to a minimum:

- *Task 1: Identification of relevant articles*: Instead of relying on the claim identifier to separate relevant from irrelevant articles, we would propose to either introduce manual screening of the claim sentences the model proposes or fine-tune a topic specific claim detection model with a small number of curated claims. This would allow, to get rid of the false positives with limited resources.

  Another option for narrowing down the set of initially found articles would be Topic Models. This was not part of our pipeline, but by utilising Topic Models, all articles are initially assigned to the most relevant topics based on their topic distribution. For each topic, a subset of articles is manually classified whether it fits into the relevant articles or not. This process generates a diverse sample, ideal for training a relevant articles classifier. The Topic Vector of each article can be utilised for the training process. This approach offers a robust and efficient way to curate a high-quality set of articles in a given issue area.
- *Task 2: Identification of sentences/text segments that contain claims:* Topic agnostic claim identification with a model trained on the original MARDY dataset, containing only claims about migration works well. Nevertheless, we expect that fine-tuning the model on a small set of curated annotations and other model improvements like actor masking (Dayanik and Padó 2020) will likely increase the F1 score some points.

- *Task 3: Identification of actors who make the claims:* To improve actor identification a real-world application would include co-reference resolution to capture also the actor-claim dyads from sentences in which only pronouns are used. But, since co-reference resolution requires context awareness beyond individual sentences, this may require switching away from sentence transformer models, which, on the other hand, perform better than their token/text-based pendants. In addition, a high-quality actor identification model would ideally link the found named entities to a knowledge base so that the model can see that e.g. “Angela Merkel”, “die Bundeskanzlerin”, and “die Regierungschefin” were during a certain period the same person. Another option may be to use large language models (LLMs) like ChatGPT or LLAMA, first experiments in this direction are promising (Barić, Papay, and Padó 2024).
- *Task 4: Classification of the found claims*: For this task, our model was confronted with two problems. First, the high number of false positive claim candidates from task 1 and 2 need to be reduced, because they will always be misclassified. Second, the few-shot learning setting had problems disentangling neighbouring claim categories. In a real-world scenario we would propose to inspect the confusion matrix, and then provide more training sentences for categories that the model initially confuses more frequently. In an iterative process, model predictions can be used to identify candidates for additional training sentences which have to be manually evaluated by humans.
- *Task 5: Determination of actor positions towards the respective claim*: The good performance with a very simplistic model suggests that few additional negation patterns may increase the model performance significantly. An iterative approach could identify sentence classes with weak performance and then add negation patterns for these classes to improve the model.

So, overall, we can see that fully automatic discourse network generation based on state-of-the-art computer linguistic language models still does not work out of the box. But each step already is able to achieve high-quality results that then can be used for human curation and thus reduce the time and resources necessary to do a discourse (network) analysis of a political debate significantly. It is still not possible to “just run a model” on a given text corpus. But computer linguistic models today are already able to fundamentally alter the research process for a discourse (network) analysis. Instead of going manually through every text, researchers may use the models to filter out most of the irrelevant parts, pre-annotate most of the relevant sentences and limit human intervention to those cases where the predicted probabilities of claim categories and actors are below a certain threshold. For more reliable models it may be necessary to start with a small set of manually annotated texts. But the bulk of annotation work will be done in the future by fine-tuned models, and human researchers can spend their time more efficiently on curation tasks.

Existing studies proposing generalizable tools based on transformer models, LLMs and other supervised machine learning approaches have mostly provided solutions for coarse-grained analysis of political discourse and have addressed low-dimensional classification problems, e.g. sentiment classification with eight sentiment categories (Widmann and Wich 2023), identification of the Narrative Policy Framework's six role categories (Wolton, Crow, and Heikkila 2022), three category natural language inference tasks (Laurer et al. 2024), or binary classification of conflict vs. non-conflict events (Hu et al. 2022). Our experiment shows that automation using transformer-based models can also address more complex problems where models have to distinguish between a much larger number of categories, and thus address a much more fine-grained classification problem.